\documentclass[runningheads]{llncs}

\usepackage[T1]{fontenc}
\usepackage{graphicx}
\usepackage{xcolor}
\usepackage{amsmath}
\usepackage{algorithm}
\usepackage{algorithmic}
\usepackage{multirow}
\usepackage{marvosym}

\begin{document}
\title{SummScore: A Comprehensive Evaluation Metric for Summary Quality Based on Cross-Encoder}
\titlerunning{SummScore}
%


\author{Wuhang Lin\inst{1}\textbf{*}, Shasha Li\inst{1}\textbf{*}, Chen Zhang\inst{1}, Bin Ji\inst{1}, Jie Yu\inst{1}\textsuperscript{\Letter}, Jun Ma\inst{1}, \and Zibo Yi\inst{2}
}

\authorrunning{Wuhang Lin et al.}

\institute{College of Computer, National University of Defense Technology, Changsha, China \\
	\email{wuhang\_lin@163.com, shashali@nudt.edu.cn, chenzhang199705@163.com, jibin@nudt.edu.cn, yj@nudt.edu.cn, majun@nudt.edu.cn} \and
	Information Research Center of Military Science PLA Academy of Military Science 100142 Beijing, China\\
	\email{ziboyi@outlook.com}
}

\maketitle              
\begin{abstract}
Text summarization models are often trained to produce summaries that meet human quality requirements.
However, the existing evaluation metrics for summary text are only rough proxies for summary quality, suffering from low correlation with human scoring and inhibition of summary diversity. 
To solve these problems, we propose SummScore, a comprehensive metric for summary quality evaluation based on Cross-Encoder.
Firstly, by adopting the original-summary measurement mode and comparing the semantics of the original text, SummScore gets rid of the inhibition of summary diversity.
With the help of the text-matching pre-training Cross-Encoder, SummScore can effectively capture the subtle differences between the semantics of summaries.
Secondly, to improve the comprehensiveness and interpretability, SummScore consists of four fine-grained submodels, which measure Coherence, Consistency, Fluency, and Relevance separately. 
We use semi-supervised multi-rounds of training to improve the performance of our model on extremely limited annotated data.
Extensive experiments show that SummScore significantly outperforms existing evaluation metrics in the above four dimensions in correlation with human scoring.
We also provide the quality evaluation results of SummScore on 16 mainstream summarization models for later research.

\end{abstract}

\section{Introduction}

Automatic text summarization technology aims to compress a long document into a fluent short text, which is consistent with the key information of the original text and preserves the most salient information in the source document~\cite{2020FEQA}. 
In recent years, automatic text summarization technologies have been significantly developed. 
However, the research on automatic summarization evaluation still fell behind~\cite{fabbri2021summeval}. 
Today, the mainstream evaluation metrics for automatic text summarization, such as ROUGE, BLEU, and Meteor, simply calculate n-gram overlap between candidates and references~\cite{banerjee2005meteor,lin2004rouge,papineni2002bleu}. 
Studies~\cite{lloret2018challenging,2020Learning} have shown that they are only rough proxies for summary quality evaluation. 
Some concerns of these metrics are shown as follows.

Firstly, the existing evaluation metrics strongly rely on expert-generated summaries as references, which are difficult to obtain. 
What’s more, these metrics inhibit the diversity of summaries generated by the summarization model. 
Because the mainstream metrics only rely on the interaction between the reference summary. 
However, different summaries written by readers with different knowledge reserves and for different purposes are also correct. 
We cannot force different summaries to be evaluated simply by measuring the degree of alignment with a single reference summary. 
Such an evaluation metric will limit the diversity of summaries generated by the summarization model.

Secondly, some studies show that the mainstream evaluation metrics scoring do not correlate well with human scoring~\cite{2021A,2020Learning}.
When humans evaluate the quality of summaries, they usually consider multiple fine-grained quality dimensions, such as rich information, non-redundancy, coherence, and well-structured.
However, these metrics mainly focus on the similarity of literal and expressions, which cannot well evaluate semantic relevance and topic consistency. 
Moreover, they ignore the evaluation of language quality, such as logical consistency and language fluency. 
Many of the above-mentioned factors can affect the comprehensiveness and interpretability of the summary quality evaluation.

\begin{figure}[h]
	\centering
	\includegraphics[width=0.8\columnwidth]{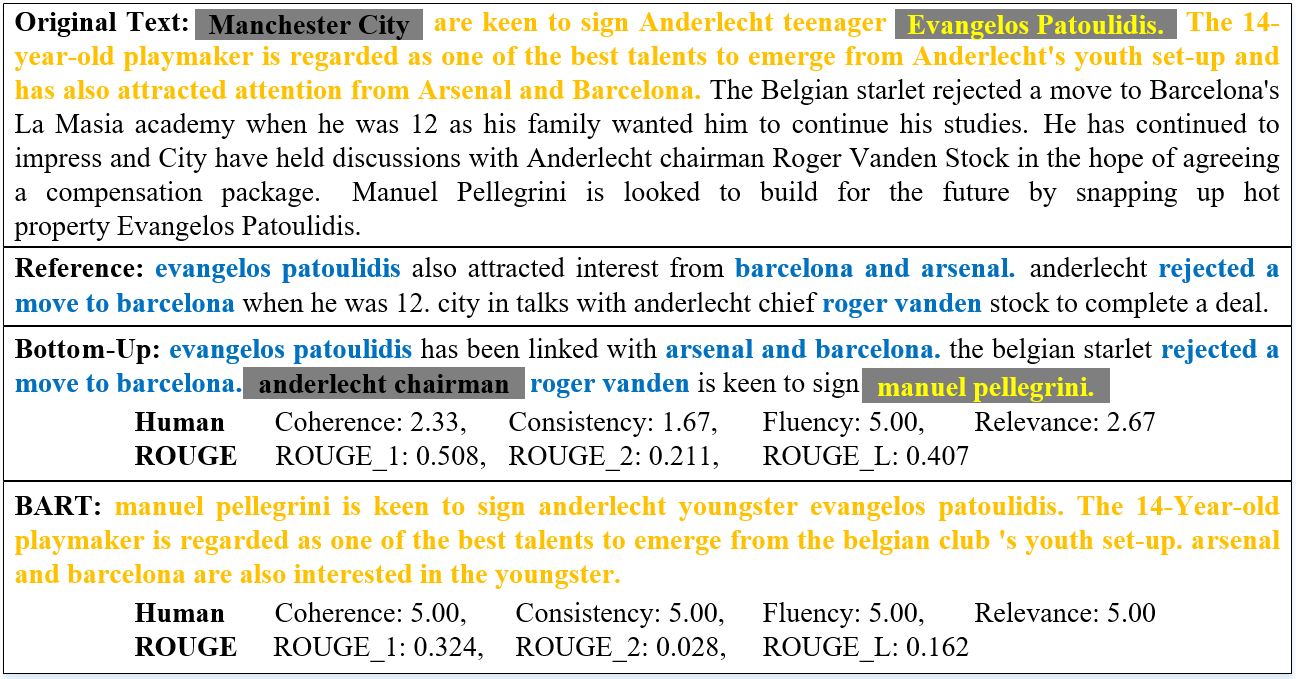}
	\caption{A typical example showing ROUGE's problems.}
	\label{figexample}
\end{figure}

As illustrated in the examples in Figure.~\ref{figexample}.
Comparing the reference with the original text, when experts score the summary generated by model Bottom-Up, they find that the generated summary has \textcolor{gray}{factual errors(gray shaded fonts)}.
The fact is that \emph{Manuel Pellegrini (Manchester City)} wants to sign \emph{Evangelos Patoulidis}.
Therefore, except for Fluency, the experts give low scores for all quality dimensions.
However, because of \textcolor{blue}{the large overlap of n-grams(blue fonts)} between the summary and the reference, ROUGE scores high.
For the BART model, because the generated summaries almost focus on \textcolor{orange}{the important information (orange fonts)}, and the text is of high quality and no redundant information. 
So, experts give it high marks.
However, the wording is different from the reference summary, so ROUGE gives the summary a low rating.
It can be seen from these two examples that ROUGE is a rough proxy that is unable to recognize semantic factual errors.
Moreover, over-reliance on the literal matching of reference may lead to a suppression of the diversity of generated summaries.
Therefore, a good summarization evaluation metric should be able to help identify:
(i) semantically correct summaries with good word overlap with the original text or reference,
and (ii) non-redundant and fluent summaries that contain enough correct facts, even if their wording is different from the reference.

To solve these problems, we propose SummScore, a comprehensive metric for summary quality evaluation based on Cross-Encoder.
SummScore adopts the original-summary paired measurement mode. 
The summaries are scored by comparing the semantics of the original text, avoiding the suppression of the diversity of the summaries caused by the forced alignment of a single reference summary. 
With the help of the text-matching pre-training Cross-Encoder, SummScore can effectively capture the subtle differences between the semantics of summaries.
To improve the comprehensiveness and interpretability, SummScore consists of four fine-grained submodels, which measure Coherence, Consistency, Fluency, and Relevance separately.

We conduct our experiment in SummEval\cite{fabbri2021summeval} dataset and measure the quality of our SummScore by calculating the Pearson correlation and Spearman correlation coefficient between SummScore scores and human annotation scores.
We use semi-supervised multi-rounds of training to improve the performance of our model on extremely limited annotated data.
Extensive experiments show that SummScore significantly outperforms existing evaluation metrics.
In addition, we evaluate 16 mainstream summarization models with SummScore and publish the results for later research.
Our contributions are summarized as follows:

\begin{itemize}
	\item We propose SummScore, a novel evaluation metric for summary quality, which uses original text instead of the hard-to-obtain expert-generated gold summary as the reference to evaluate the quality of the generated summary.
	
	\item We trained four submodels of SummScore based on the Cross-Encoder framework to automatically evaluate the four fine-grained qualities of Relevance, Consistency, Fluency, and Coherence respectively. 
	Experiments show SummScore has strong human relevance on all the four fine-grained dimensions.
	
	\item We evaluate 16 mainstream summarization models with SummScore and publish the results for later research.
\end{itemize}

\section{Related Work}

In this section, we will first introduce the common metrics on summarization and their main problems.
Next, we will introduce the context-dependent metrics and the trained metrics in the evaluation of related natural language generation tasks.
By borrowing the principles and advantages of the context-dependent metrics and the trained metrics, we design SummScore for summary quality evaluation.

\subsubsection{Common Metrics in Summarization}
The early common summary metrics are mainly represented by ROUGE~\cite{lin2004rouge}, BLEU~\cite{papineni2002bleu} and METEOR~\cite{banerjee2005meteor}.
All of them obtain the summary quality score by calculating the token n-gram overlap between the summary and the reference.
However, these lexical-based overlap metrics cannot capture the changes in semantics and grammar.
Therefore, BERTscore~\cite{zhang2019bertscore} and MoverScore~\cite{zhao2019moverscore} use BERT to extract contextual embeddings and use embeddings matching to complete the similarity calculation between summary and references. 
However, these metrics, which rely on the alignment of single-reference abstracts, bring about suppression of abstract diversity.

\subsubsection{Context-dependent Metrics}
To get rid of the constraints of reference summaries, ROUGE-C~\cite{he2008rouge} improves ROUGE, which compares summaries with the original texts instead of reference summaries. ROUGE-C proves that using original text instead of reference can yield positive benefits, especially when the reference summary is not available.
SUPERT~\cite{gao2020supert} is an unsupervised reference-less summarization evaluation metric.
SUPERT enables the quality assessment of the generated summaries with the help of pseudo-reference summaries created by salient sentences from the original text. 
Our model is also a context-dependent metric.
Experiments show that our method not only gets rid of the comfort of reference summary but also supports the diversity of summary text generation.

\subsubsection{Trained Metrics}
There are training-based evaluation metrics in related natural language generation tasks.
For machine translation, BLEND~\cite{ma2017blend} and BEER~\cite{stanojevic2014fitting} train a scoring model by combining a variety of existing untrained metrics, such as BLEU, METEOR, and ROUGE.
As the pre-training models show promising performance, BERT for MTE~\cite{shimanaka2019machine} and BLEURT~\cite{sellam2020bleurt} are proposed for the machine translation system.
By performing BERT fine-tuning training on a small amount of labeled data, they compute the similarity of the candidate and reference sentences.
The difference is that BLEURT innovatively designs a set of pre-training signals and pre-trains BERT.
We propose a trained-based summary evaluation metric SummScore, which consists of four submodels, corresponding to four quality dimensions.
We believe that a single model may not be able to take into account the evaluation of various quality dimensions of the summary texts.
At the same time, the independent scoring of multiple dimensions also helps to improve the interpretability of the summary quality score.

\section{Our Methodology}

\subsection{Problem Definition}
Our SummScore model is based on the \textbf{Cross-Encoder}~\cite{2020RocketQA} model in the field of information retrieval.
In QA(Question answering) retrieval, when sorting the candidate answers, the higher the similarity score between the answer and the question, the more accurate the answer is considered.
The specific process can be realized by stitching the subword sequences of question text and answer text with [SEP] and inputting them into the Cross-Encoder model for training.
Similarly, a summary can be regarded as a semantically similar text obtained after the original text is compressed.
A heuristic idea is that the more similar the summary is to the original text, the higher the quality of the summary. 
The similarity here includes semantic similarity, content consistency, etc.
Inspired by QA retrieval, we also regard the scoring of summary quality as a process of text similarity calculation between the original text and summary text.

\begin{figure*}[t]
	\centering
	\includegraphics[width=0.9\textwidth]{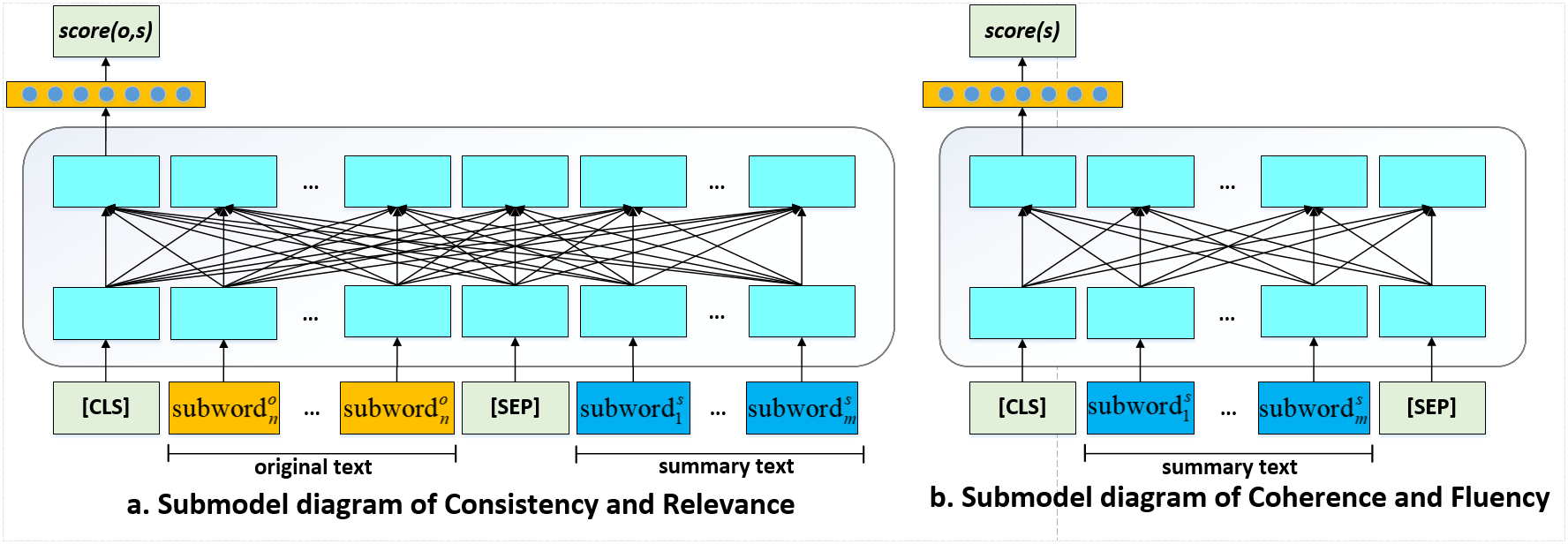} 
	\caption{Structural diagram of SummScore's submodels.}
	\label{cross_encoder}
\end{figure*}

As shown in Figure.~\ref{cross_encoder}, we formally define the summary quality evaluation problem as follows.
Given the subword sequence $O$ of the source document, where $O=\{o_{1},...,o_{n}\}$.
Suppose that the subword sequence of the generated summary is $S$, where $S=\{s_{1},...,s_{m}\}$.
The goal is to implement a function $score(O, S)$ and predict a score $y$ to represent the similarity between document $O$ and summary $S$.
Given the training data with human annotation scores on summary quality, our goal is to train the function $score(O, S)$ so that it can regress to the human annotation score $y^{'}$.

\subsection{Structure of Model}
The structure of SummScore's submodels is based on the Cross-Encoder.
The Cross-Encoder~\cite{2019Sentence} believes that the spliced sentence pair is a reasonable input mode, which is suitable for NSP(Next Sentence Prediction)~\cite{devlin2018bert} pre-training task and natural language inference task.
Our SummScore is designed based on the principle of semantic similarity computation, and the used Cross-Encoder is pre-trained on related tasks.
Hugging Face SentenceTransformers provides researchers with Cross-Encoder\footnote{https://www.sbert.net/examples/training/cross-encoder/README.html} after training on the semantic similarity benchmark dataset STS~\cite{2017SemEval}.
By inputting sentence pairs, the Cross-Encoder will predict a score between 0 and 5 representing the semantic similarity of the two texts.

Subsequently, we use the pre-trained Cross-Encoder to perform fine-tuning on fine-grained quality human-annotated data.
Specifically, we add a regression model based on MLP(Multilayer Perceptron) to Cross-Encoder to evaluate the score.
The format of the input is a patchwork of sentence pairs.
The first token of each sentence pair is always a special mark [CLS], and the sentences are separated by [SEP].
Finally, the final hidden state corresponding to the first special [CLS] token is taken as the sentence feature of the overall input.
Feed the [CLS] embedding $V_{[CLS]}$ into the MLP to get the predicted score $y$:
\begin{alignat}{2}
	V_{[CLS]}&=Cross-Encoder([CLS],O,[SEP],S) \label{cls}\\
	y &= WV_{[CLS]}+b 
\end{alignat}
where $W$ and $b$ are learnable parameters.
The learning goal of the whole model is to fit the gold label $y'$ with $y$.
Our squared regression loss is:
\begin{alignat}{2}
	\mathcal{L}&=\frac{1}{N} \sum_{n=1}^{N} \lVert y-y' \rVert ^{2} 
\end{alignat}
where $N$ is the size of the sample.

\subsection{Training Method of Submodels}
Proxy metrics such as ROUGE and BERTscore usually return only a single value for the summary quality.
It is difficult for people to clearly know how good or bad the current summary is from this score value.
For example, does this summary capture the topic of the original text? How fluent is this summary? What are the main problems in this summary?
That is, proxy metrics such as ROUGE and BERTscore are not well interpretable.
Due to the poor interpretability of metrics scoring, it is also difficult for the summarization model to further improve the quality of the generated summary and the performance of the model.

The counterpart to machine scoring is human evaluation.
It is a common fact that the human evaluation will first divide the quality of summaries into multiple fine-grained quality dimensions, and then score on the specific dimensions.
A popular division is to divide the quality of the summary into four fine-grained quality dimensions (\textbf{Coherence}, \textbf{Consistency}, \textbf{Fluency}, and \textbf{Relevance})~\cite{fabbri2021summeval,W2019Neural}.
Specifically, \textbf{Coherence}: the summary should be a coherent set of information about a topic, and whether the organizational structure between sentences is logical. 
\textbf{Consistency}: the summary should contain only the facts and themes of the original text. Both should be presented consistently and without hallucinatory facts.
\textbf{Fluency}: the quality of the language. Whether there are grammatical errors that affect reading.
\textbf{Relevance}: the summary should only contain important information from the source document, penalizing the summary that contains redundant information.
Following this principle, our SummScore is also composed of four scoring submodels, and each corresponds to one of the above quality dimensions.
Therefore, SummScore has the human-like ability to comprehensively evaluate the quality of summaries across multiple quality dimensions.

\begin{algorithm}[tb]
	\caption{The semi-supervised multi-round training}
	\label{train_algorithm}
	\textbf{Input}: Initial submodel $M_{0}$, annotated dataset $D_{L}(D_{L}^{train}\cup D_{L}^{val})$, unannotated data subset $D_{U}=\{D_{1},...,D_{k}\}$, epoch size for fine-tuning $ep$\\
	\textbf{Output}: The best submodel $M_{best}$ 
	\begin{algorithmic}[1] 
		\STATE  /*Part1: the first round of supervised training with $D_{L}^{train}$.*/ 
		\STATE Let $M_{0}^{best}=M_{0}$\;  
		\FOR{each $i\in \{0,1,...,ep-1\}$} 
		\STATE Train $M_{0}$ on $D_{L}^{train}$ an epoch agin and obtain $M_{i+1}$\;
		\IF {$f(M_{i+1},D_{L}^{val})>f(M_{best},D_{L}^{val})$}
		\STATE $M_{0}^{best}=M_{i+1}$\;
		\ENDIF
		\ENDFOR \\\
		\STATE  /*Part2: multiple rounds of semi-supervised training. */ 
		\STATE Let $D=D_{L}^{train}, M^{best}=M_{0}^{best}$\;
		\FOR{each $t\in \{1,2,...,k\}$} 
		\STATE Annotate $D_{t}$ with $M_{t-1}^{best}$ and obtain pseudo-annotated data $D^{pseudo}_{t}$
		\STATE $D=D \cup D^{pseudo}_{t}$
		\STATE Let $M_{t}^{best}=M_{0}$\;
		\FOR{each $i\in \{0,1,...,ep-1\}$}
		\STATE Train $M_{i}$ on $D_{L}^{train}$ an epoch agin and obtain $M_{i+1}$ \;
		\IF {$f(M_{i+1},D_{L}^{val})>f(M_{t}^{best},D_{L}^{val})$}
		\STATE $M_{t}^{best}=M_{i+1}$\;
		\ENDIF
		\ENDFOR
		\IF {$f(M_{t}^{best},D_{L}^{val})>f(M^{best},D_{L}^{val})$}
		\STATE $M^{best}=M_{t}^{best}$\;
		\ENDIF
		\ENDFOR
	\end{algorithmic}
\end{algorithm}

The model structures of the four submodels are consistent, but the mode of data input of the submodels of Coherence and Fluency is different.
As shown in Figure.~\ref{cross_encoder}, among them, the scoring submodels for Fluency and Coherence no longer use the training mode of sentence pair.
Because Fluency evaluates the linguistic quality of the summary itself.
When experts annotate Fluency's scores, they can do it without referring to the original text.
For the Coherence dimension, experts only focus on whether the summary text itself has a clear theme and rigorous sentence logic.
In contrast, when experts score the quality dimensions of Consistency and Relevance, it is necessary to repeatedly compare the generated summary with the original text.
Therefore, for the submodels of Coherence and Fluency dimension, we remove the original text information and change the formula \eqref{cls} to the following form:
\begin{alignat}{2}
	V_{[CLS]}&=Cross-Encoder([CLS],S,[SEP])
\end{alignat}

Because the annotation data resources are very limited, we adopt a semi-supervised multi-round training method to maximize the correlation between SummScore and human ratings.
The input of the algorithm includes the pre-trained Cross-Encoder $M_{0}$, which is used as the initial state of the SummScore's submodel.
We have a small-scale manually annotated supervised dataset $D_{L}$.
We divide $D_{L}$ into the training set $D_{L}^{train}$ and validation set $D_{L}^{val}$.
In addition, we have a large amount of unsupervised data $D_{U}$ generated by several mainstream summarization models.
$D_{U}$ is randomly divided into sub-datasets of the same size $\{D_{1},...,D_{k}\}$.
Moreover, we also have a scoring function $f($·$)$ to judge whether the submodel is good or bad, which is achieved by comparing the correlation between the scores predicted by the submodel and the manually annotated scores on $D_{L}^{val}$.
$f($·$)$ can be chosen from $max(Pearson)$, $max(Spearma)$ and $max(Pearson * Spearma)$.
Our goal is to obtain the globally optimal submodel $M_{best}$ with limited annotation data.

Our training is mainly divided into two parts, as shown in lines 1-8 and 9-24 of the Algorithm.~\ref{train_algorithm} respectively.
In the first part of the algorithm, we first train the submodel on the small-scale supervised data $D_{L}^{train}$.
In the beginning, we assume that the best submodel $M_{0}^{best}$ in the first round of training is $M_{0}$ (line 2).
Then, we perform fine-tuning training for $ep$ times (line 3).
After the $i$-th fine-tuning, the submodel is trained from $M_{i}$ to $M_{i+1}$ (line 4).
After each fine-tuning, we compare the quality of $M_{0}^{best}$ and $M_{i+1}$, and save the best model as $M_{0}^{best}$ (line 5-7).
After the first round of supervised training, we get the best model of the first round $M_{0}^{best}$.

In the second part of the algorithm, we will carry out multiple rounds of semi-supervised training to improve the performance of the submodel using unlabeled data.
We assume that the initial global optimal model is $M_{0}^{best}$, and the current training available dataset $D$ is $D_{L}^{train}$ (line 10).
Because the unsupervised dataset $D_{U}$ is divided into $k$ blocks, the algorithm will perform $k$ rounds of semi-supervised training (Line 11).
At the beginning of the $t$-th round of semi-supervised training, we will use the optimal model of the previous round $M_{t-1}^{best}$ to label the sub-data $D_{t}$, and get the pseudo-labeled dataset $D^{pseudo}_{t}$ (line 12).
Then, the newly obtained pseudo-label data $D^{pseudo}_{t}$ is extended to the available dataset $D$ for the next round of semi-supervised training (line 13).
After that, like the steps of Part1, start with the initial Cross-Encoder $M0$ and fine-tune the submodel $ep$ epochs on data $D$ (line 14-20).
After the end of each epoch fine-tuning, the optimal submodel $M_{t}^{best}$ of round $t$ is retained (line 18).
After each $t$-th round of semi-supervised training, we will also compare $M^{best}$ and $M_{t}^{best}$, and keep the best model as the global optimal model $M^{best}$(line 22).
After all $t$ rounds of semi-supervised training, we finally obtain the globally optimal submodel $M^{best}$ of SummScore for each fine-grained quality dimension.

\section{Experiments Settings}
We conduct experiments on SummEval~\cite{fabbri2021summeval} dataset, which contains 1600 manually annotated summaries. 
Each summary is evaluated on the four fine-grained quality dimensions according to criteria~\cite{W2019Neural} and is scored by 5 independent crowdsource workers and 3 independent experts.
Annotation scores span from 1 (worst) to 5 (best).
We calculate the average of the annotation scores of the 3 experts as the final supervision score for each data and randomly divide the data into a training set $D_{L}^{train}$ of 1000 pieces of data and a test set $D_{L}^{test}$ of 600 pieces of data. 
At the beginning of each round of semi-supervised training, we randomly sample 100 pieces of data from the training set $D_{L}^{train}$ as the validation set $D_{L}^{val}$ for model selection and pass it to $f($·$)$ for model selection.

In addition to the above small-scale annotated data, we also use a large amount of unannotated data consisting of original texts and machine-generated summaries.
These unannotated data will be randomly divided into $k$ equal-sized parts in the experiment.
Specifically, these divided data are mainly used in the semi-supervised training of the model to further help SummScore improve performance.

Our SummScore is based on Cross-Encoder\footnote{https://www.sbert.net/examples/training/cross-encoder/README.html} of Hugging Face SentenceTransformers.
We expect that the scoring process of SummScore will be as fast as possible without taking up too much video memory of the machine.
Therefore, we abandon the pre-training model with large-scale parameters, such as \emph{RoBERTa$_{LARGE}$}(24 layers), and only select the public \emph{DistilRoBERTa$_{BASE}$}(6 layers), and \emph{RoBERTa$_{BASE}$}(12 layers) for fine-tuning.
So the GeForce GTX 1060 can meet all the experimental needs of SummScore. We set the epoch size for fine-tuning to be 6 and the batch size to be 4.
When the amount of newly expanded pseudo-annotated data reaches 10,000 (about ten times the annotated data), the model can obtain satisfactory performance.
At the beginning of each new round of semi-supervised training, SummScore will perform linner warmup training with 1/10 of the single round steps.
We use Adam as our optimizer with a learning rate of 2e-5 and a weight decay of 0.01.
Consistent with previous research works~\cite{shimanaka2019machine,zhang2019bertscore}, we use Pearson and Spearman correlation coefficients to judge the correlation between the scoring metrics and manual scoring.

\section{Experiments}

\subsection{Comparative Experiments}

For the convenience of comparison, we conduct our comparative experiments in groups.
First, we compare our model with several well-known training-free metrics.
These metrics include BLEU~\cite{papineni2002bleu}, TF-IDF, ROUGE, BERTscore, and SUPERT.
These metrics have their innovative principles and advantages, which have a profound impact on the development of the corresponding field.
In particular, ROUGE and BERTscore are very popular and well-received in summary quality evaluation.
Our SummScore is based on pre-training fine-tuning.
Therefore, we also select two representative metrics based on the pre-trained model fine-tuning: BLEURT and BERT for MTE.
For a fair comparison, we maintain the experimental design consistent with SummScore and conduct fine-tuning on the same data.
Similarly, BLEURT and BERT for MTE also adopt multi-round semi-supervised training to eliminate the influence of training methods.

\begin{table*}[t] \centering
	\centering
	\caption{The results of the correlation experiment of the evaluation metrics on the test set of SummEval}
	\resizebox{\textwidth}{!}{
		\begin{tabular}{p{0.25cm}|l|cc|cc|cc|cc} \hline
			\multicolumn{2}{c|}{\textbf{Quality Dimension}} & \multicolumn{2}{|c|}{\textbf{Coherence}} & \multicolumn{2}{|c|}{\textbf{Consistency}} & \multicolumn{2}{|c|}{\textbf{Fluency}} & \multicolumn{2}{|c}{\textbf{Relevance}}  \\ \hline
			\multicolumn{2}{c|}{\textbf{Metric}}   & \small \textbf{Pearson} & \small \textbf{Spearma} & \small \textbf{Pearson} & \small \textbf{Spearma}   & \small \textbf{Pearson} & \small \textbf{Spearma} & \small \textbf{Pearson} & \small \textbf{Spearma}  \\ \hline
			\multirow{13}*{\rotatebox{90}{ Training-free}} &BLEU-1  & 0.0278  &0.0272  & 0.2023  &0.1552   &0.1367   &0.0696  &0.2459  &0.1992     \\ 
			&BLEU-2  & 0.0419  &0.0384  & 0.1531  &0.1456   &0.1206   &0.0810  &0.2104  &0.2002     \\ 
			&BLEU-3  & 0.0588  &0.0668  & 0.1129  &0.1367   &0.1092   &0.0841  &0.1901  &0.2240     \\ 
			&BLEU-4  & 0.0513  &0.0764  & 0.0896  &0.1271   &0.1053   &0.0898  &0.1567  &0.2193     \\ 
			&TF-IDF  & 0.0667  &0.0689  & 0.0772  &0.0930   &0.0797   &0.0727  &0.0893  &0.0472     \\ 
			&ROUGE-1  & 0.1757  &0.1593  & 0.2242  &0.1767   &0.1651   &0.1055  &0.3264  &0.2955     \\ 
			&ROUGE-2  & 0.1229  &0.1237  & 0.1515  &0.1489   &0.1297   &0.1062  &0.2502  &0.2509     \\ 
			&ROUGE-3  & 0.1219  &0.1339  & 0.1184  &0.1319   &0.1182   &0.1106  &0.2160  &0.2472     \\ 
			&ROUGE-L  & 0.1569  &0.1457  & 0.1671  &0.1486   &0.1578   &0.1499  &0.2415  &0.2284     \\ 
			&BERTscore-r  & 0.1414  &0.1297  & 0.2456  &0.1921   &0.2142   &0.1553  &0.3474  &0.2960     \\ 
			&BERTscore-p  & 0.1746  &0.1428  & 0.1059  &0.0821   &0.2414   &0.1683  &0.1930  &0.1513     \\ 
			&BERTscore-f  & 0.1792  &0.1534  & 0.2043  &0.1750   &0.2614   &0.1811  &0.3135  &0.2694     \\ 
			&SUPERT & 0.2853  &0.2599  & 0.3230  &0.2931   &0.3062   &0.2280  &0.3703  &0.3256     \\ \hline
			\multirow{2}*{\small \rotatebox{90}{ Trained}} &BLEURT & 0.4631  &0.4410  & 0.3206  &0.2233   &0.4639   &0.2193  &0.5621  &0.5286     \\ 
			&\small BERT for MTE  &0.5532  &0.5324       &0.3721  &0.3058      &0.4601  &0.2645     &0.5638 &0.5315      \\  
			&\small BERT for MTE$_{DistilRobertaBase}$  & 0.6080  &0.6036       & 0.4630 &0.3512       &0.4787  &0.3509     &0.5813  &0.5500      \\  \hline 
			\multirow{2}*{\rotatebox{90}{Ours}}&\small SummScore$_{DistilRobertaBase}$  & \textbf{0.6704}  &\textbf{0.6684}    &\textbf{0.4839}  &\textbf{0.4080}   &\textbf{0.7071}   &\textbf{0.5586}  &\textbf{0.6018}  &\textbf{0.5538} \\ 
			&\small SummScore$_{RobertaBase}$  & \textbf{0.7061}  & \textbf{0.7116}    &\textbf{0.4852}  &\textbf{0.4497}    &\textbf{0.7348}   &\textbf{0.5855}  &\textbf{0.6746}  &\textbf{0.6391} \\ 
			\hline
		\end{tabular}
	}
	\label{result1}
\end{table*}

Table.\ref{result1} shows our experimental results.
Scores represent the Pearson correlation and Spearman correlation of each metric with respect to human annotations.
It can be seen that compared with the training-free metrics, SummScore far exceeds them.
Moreover, except for SUPERT, these metrics have a low correlation with human annotations in all fine-grained dimensions.
However, we find that they (e.g. BLEU, ROUGE, and SUPERT) tend to be more relevant to human judgments than Coherence and Fluency on Relevance and Consistency quality dimensions.
The reason is that these metrics all need to compare the literal n-gram or semantic information of the reference summary when scoring.
Because the reference summary is a compressed text that captures the central idea of the original text.
Therefore, these metrics can achieve the purpose of preliminarily measuring the Relevance and Consistency of the original text topic semantics of the generated summary.
Unfortunately, they are designed without considering the quality requirements of Coherence and Fluency.
So these metrics tend to work poorly in the Coherence and Fluency dimensions.

Compared with the trained metric group, our model also outperforms all of them.
However, we find that these metrics also perform well after multiple rounds of semi-supervised training on data.
To eliminate the influence of the pre-trained language model, we also replace the pre-trained model of BERT for MTE with the same \emph{DistilRobertaBase} trained on the STS dataset.
We find that the performance of BERT for MTE model is more competitive.
This proves that the idea of SummScore's quality evaluation design, which is inspired by the similarity matching principle of information retrieval, is reasonable.

\begin{figure}[h]
	\centering
	\includegraphics[width=0.5\columnwidth]{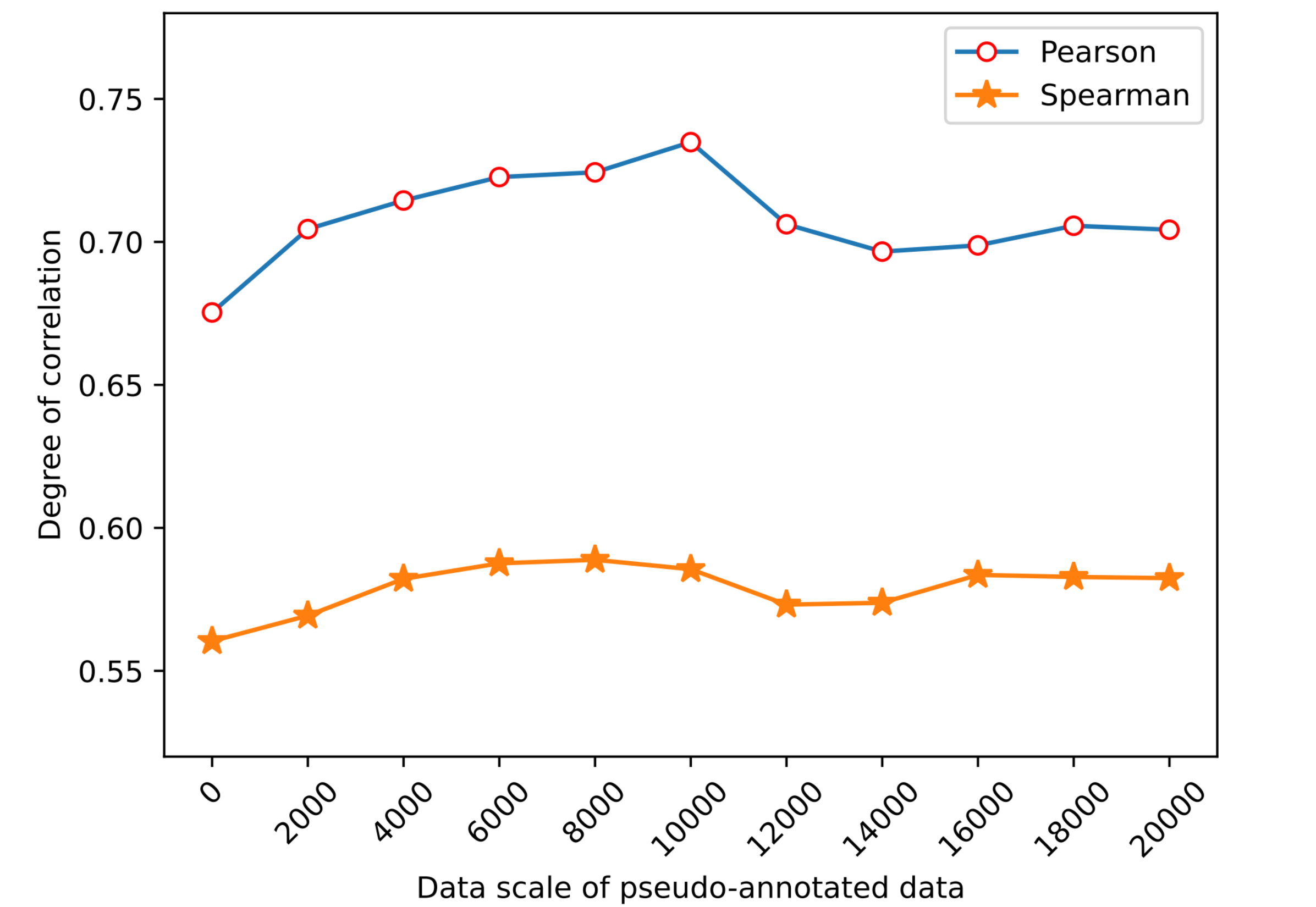}
	\caption{Ablation experiments on the impact of pseudo-annotated data volume on the Fluency dimension of \emph{SummScore$_{RobertaBase}$}.}
	\label{figdata_ablation}
\end{figure}

\subsection{Ablation Results}
We believe that multi-round semi-supervised training is an important factor for improving SummScore.
Because this training method brings about the rapid expansion of pseudo-annotated data and alleviates the problem of the small amount of data.
In order to explore the influence of the amount of pseudo-annotated data, we conduct corresponding ablation experiments.
Only the ablation experiments of \emph{SummScore$_{RobertaBase}$} on the Fluency quality dimension are introduced here, and other quality dimensions have the same conclusion.
In the ablation experiments, we expand the pseudo-annotation data with a span of 2000 pieces per round.
The results of the ablation experiment are shown in Figure.~\ref{figdata_ablation}.
It can be clearly found that the performance of the model is significantly improved with the increase of pseudo-annotation data in the early stage.
This indicates that the scale of data volume is an important factor to limit the model performance during this period.
When the expanded pseudo-annotation data reaches about 10,000 (10 times the annotation data), the correlation of Fluency reaches its peak.
This shows that at this time, the model has maximized the benefits from the increase in data volume.
Subsequently, even with more data, the performance of the model does not improve and even begins to degrade.
The ablation experiments show that reasonable multi-round semi-supervised training can effectively improve the performance of the SummScore in the case of scarce annotated data.
This also provides a new training idea for later researchers to alleviate the limitation of small data volume in similar experimental scenes.

To explore the difference between original-summary pairing $[O||S]$ and common summary-reference $[S||ref]$ (adopted by BLEURT, BERT for MTE, and other models), we also conduct relevant analysis experiments.
Table.~\ref{result3} shows the correlation of the two input methods on the \emph{SummScore$_{RobertaBase}$} model with human evaluation in the dimensions of Consistency and Relevance, respectively.
We can find that $[O||S]$ can achieve better results than $[S||ref]$.
Originally, we are also worried that the longer original information in $[O||S]$ may be more difficult for SummScore to learn than the short reference summary in $[S||ref]$.
Analyzing the reasons for the better results of $[O||S]$, we believe that the reason is that the form of $[O||S]$ may be more consistent with the scoring process of humans in the dimensions of Consistency and Relevance.
Because, normally, humans start to write a summary after reading the original text.
In real life, few golden summaries can be repeatedly referred to write the new summaries.
In the scoring process, experts often score only after reading the original text.
The input mode $[O||S]$ is also consistent with the expert scoring process.

\begin{table*}[t]
	\centering
	\caption{Ablation experiments on the influence of input modes $[S||ref]$ and $[O||S]$ on Consistency and Relevance.}
	\begin{tabular}{l|cc}
		\hline
		&  \small \textbf{Pearson} & \small \textbf{Spearma} \\
		\hline
		\textbf{Consistency$_{[S||ref]}$}    &0.4291   & 0.3290     \\
		\textbf{Consistency$_{[O||S]}$}   & \textbf{0.4852}  & \textbf{0.4497}     \\ 
		\hline
		\textbf{Relevance$_{[S||ref]}$}       & 0.6519  &0.6172    \\
		\textbf{Relevance$_{[O||S]}$}          & \textbf{0.6746}  & \textbf{0.6391}      \\
		\hline
	\end{tabular}
	\label{result3}
\end{table*}

\begin{figure}[t]
	\centering
	\includegraphics[width=0.45\columnwidth]{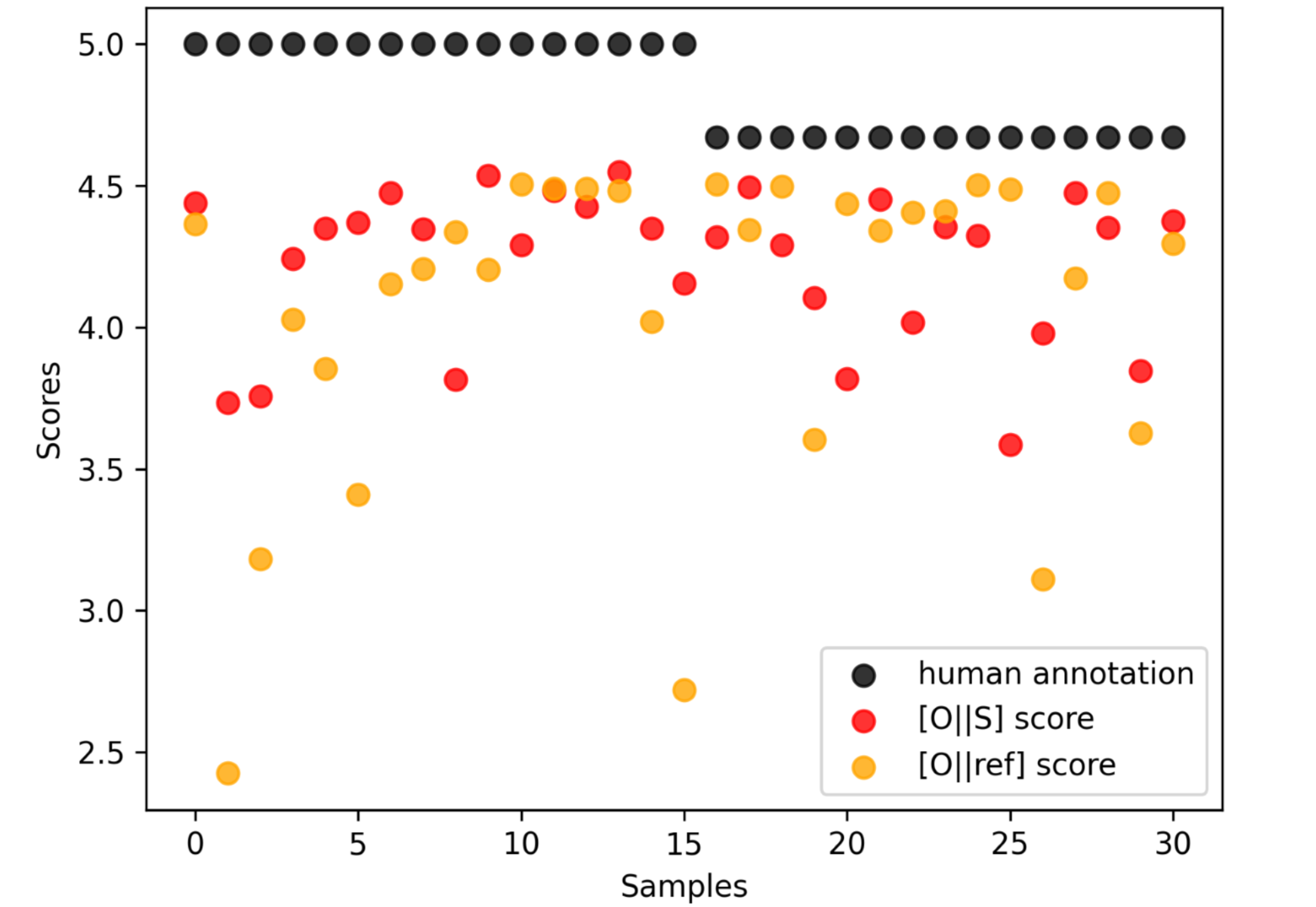}
	\caption{Ablation experiments exploring the effect of input modes $[S||ref]$ and $[O||S]$ on the diversity of summaries.}
	\label{diversity_ablation}
\end{figure}

Further experiments, we find that the original-summary mode $[O||S]$ can support the diversity of textual representations of summaries.
The lower the ROUGE score, the more different the expression of the summary and the reference.
However, those semantically correct summaries, which are expressed differently from the reference summaries, are also qualified summaries.
Qualified summary metrics should be able to identify summaries that are diverse in expression but of acceptable quality.
From the SummEval annotation data, we extract summary data with a low ROUGE score but a high human score.
We plot the scatter plots of human scores and SummScore scores for these two input modes, respectively.
Only the experiments in the Relevance dimension are listed here, and the results are shown in Figure.~\ref{diversity_ablation}.
We can find that $[O||S]$ is closer to the distribution of human scoring.
However, for the summaries with high human scores, $[S||ref]$ is more likely to give low scores.
Therefore, this can be illustrated that $[O||S]$ can recognize summaries with different literal expressions but qualified quality.
This also shows that the $[O||S]$ mode can help to improve the diversity of summary generation.

\subsection{Case Analysis}

\begin{figure}[t]
	\centering
	\includegraphics[width=0.8\columnwidth]{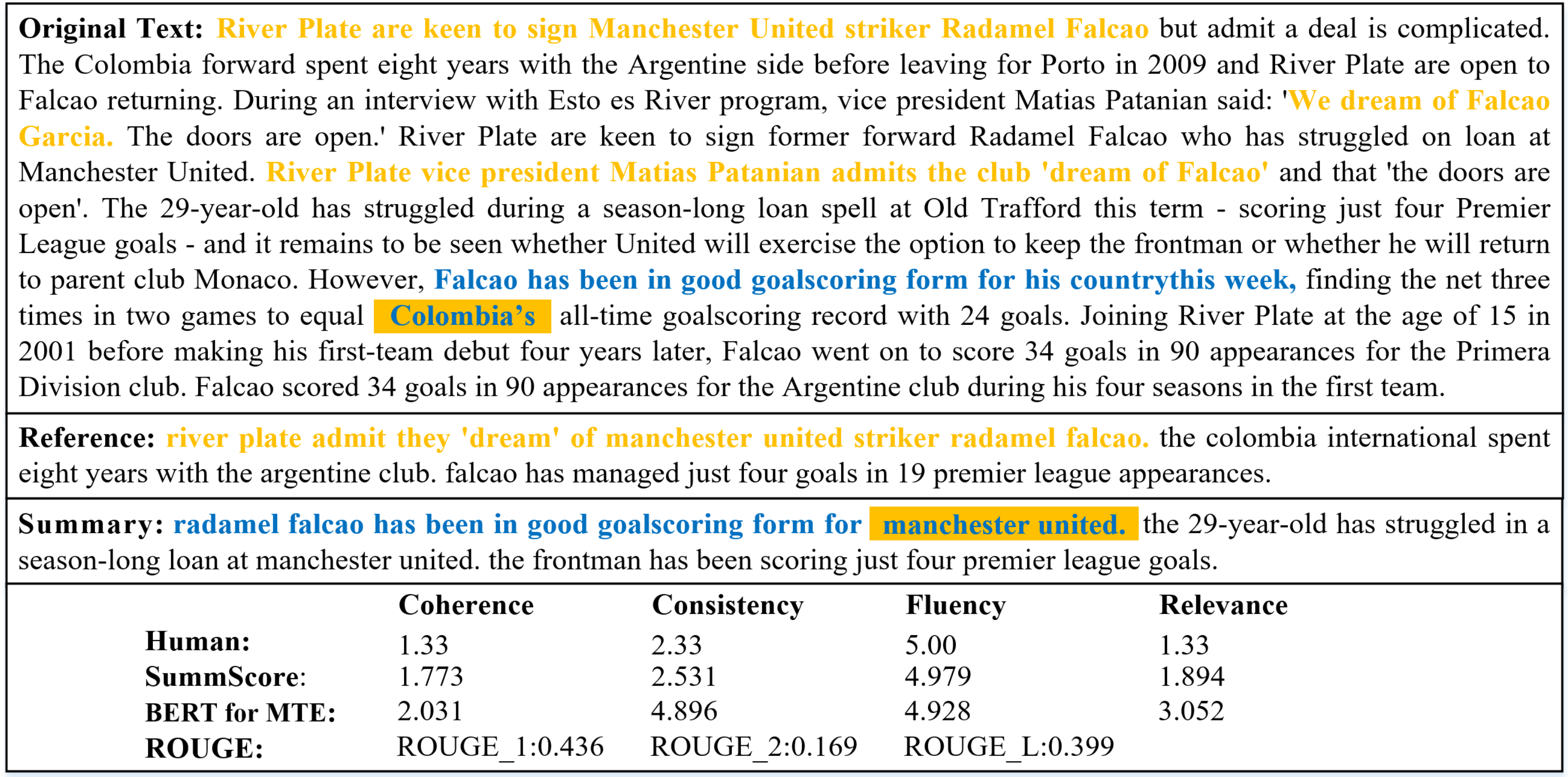}
	\caption{A classic example of performance comparison between SummScore and other metrics.}
	\label{case2}
\end{figure}

In Figure.~\ref{case2}, we show a typical example for case analysis.
By reading the original text and the reference summary, we know that the original text is about \textcolor{orange}{\emph{River Plate are keen to sign Manchester United striker Radamel Falcao}(orange fonts)}, and then some information about Radamel Falcao is introduced.
However, we can find that the summary under test completely fails to capture the central idea of the original text.
Therefore, in the Relevance dimension, both SummScore and experts give a low score of less than 2 points, but the baseline BERT for MTE scores a qualified score of 3 points.
Further analysis, we find that there are \textcolor{blue}{hallucination errors (blue shaded text)} in the summary under test.
Radamel Falcao has good goalscoring form in \emph{Colombia} rather than \emph{Manchester United}.
So both SummScore and experts give low marks for Consistency.
Analyzing the structure between sentences, we find that the semantics of the summary to be tested is lack logic.
For one moment, the summary tells us Radamel Falcao has good goalscoring form and another point that he struggles at Manchester United.
Due to the lack of logic between sentences, it is difficult to read.
So both SummScore and experts score low on the Coherence dimension, but BERT for MTE scores a high score close to 5.
In terms of fine-grained quality dimension, it can be said that SummScore has better scoring ability than BERT for MTE, and the scoring effect is closer to human scoring.
Because of the good n-gram overlap between the summary and the reference, ROUGE-1 gave this incomplete summary a high score of 0.436.
As you can see, ROUGE is indeed a rough proxy indicator without explanatory power.
ROUGE cannot tell us the specific quality of the summary, such as whether factual errors and grammatical errors exist.

\begin{table*}[h] \centering
	\centering
	\caption{Evaluation results of mainstream models on the SummScore indicator on the CNN/DailyMail.}
	\resizebox{\textwidth}{!}{
		\begin{tabular}{l|c|c|c|c|c|c|c} \hline
			\textbf{Metrics} & \textbf{ROUGE-1}& \textbf{ROUGE-2} &\textbf{ROUGE-L} & \textbf{Coherence} & \textbf{Consistency} & \textbf{Fluency} & \textbf{Relevance}  \\ \hline
			\multicolumn{8}{c}{\textbf{Extractive Models}}   \\ \hline
			LEAD-3      &0.3994  &0.1746 &0.3606 &\textbf{3.9146}  &\textbf{4.9766}  &\textbf{4.9430}  &\textbf{4.4264}    \\
			NEUSUM    &\textbf{0.4130} &\textbf{0.1893} &0.3768  &3.1327  &\textbf{4.9712}  &\textbf{4.9031}  &4.1876    \\  
			BanditSum  &\textbf{0.4137} &0.1868 &0.3759 &3.2399  &\textbf{4.9738}  &\textbf{4.9139}  &4.2007    \\ 
			RNES     &0.4088 &\textbf{0.1878} &0.3719   &\textbf{3.7673} &\textbf{4.9771}  &\textbf{4.9041}  &\textbf{4.4521}    \\  \hline
			\multicolumn{8}{c}{\textbf{Abstractive Models}}   \\ \hline
			Pointer Generator &0.3921 &0.1723 &0.3599 &3.3892  &4.9654  &\textbf{4.9401}  &4.1721    \\  
			Fast-abs-rl   &0.4057 &0.1774 &\textbf{0.3806}  &2.2031  &4.9255  &4.6389  &3.9024    \\ 
			Bottom-Up   &0.4124 &0.1870 &\textbf{0.3815}  &2.8551  &4.9113  &4.7716  &3.8890     \\ 
			Improve-abs  &0.3985 &0.1720 &0.3730   &2.1961  &4.8243  &4.5633  &3.6758     \\ 
			Unified-ext-abs   &0.4038 &0.1790 &0.3675 &3.4100  &\textbf{4.9736}  &4.8955  &4.2684     \\ 
			ROUGESa &0.4016 &0.1797 &0.3679 &3.2674  &4.9700  &4.8688  &4.1793     \\ 
			Multi-task (Ent + QG ) &0.3952 &0.1758 &0.3625 &3.3573  &4.9633  &4.8870  &4.1208     \\
			Closed book decoder &0.3976 &0.1760 &0.3636     &3.3825  &4.9688  &4.8908  &4.1866     \\
			T5    &\textbf{0.4479} &\textbf{0.2205} &\textbf{0.4172} &3.6991  &4.9126  &4.8703  &\textbf{4.3365}     \\
			GPT-2 (supervised) &0.3981 &0.1758 &0.3674  &\textbf{3.7410}  &3.9252  &3.8176  &3.6069     \\
			BART &\textbf{0.4416} &\textbf{0.2128} &\textbf{0.4100} &\textbf{4.2064}  &4.9707  &4.8545  &\textbf{4.5683}     \\
			Pegasus  &\textbf{0.4408} &\textbf{0.2147} &\textbf{0.4103} &\textbf{3.7148}  &4.9176  &4.8522  &\textbf{4.3421}     \\ 
			\hline
		\end{tabular}
	}
	\label{result5}
\end{table*}

\subsection{Mainstream Summarization Models Evaluation}

Finally, we use SummScore to evaluate the performance of 16 mainstream summarization models on the CNN/DailyMail dataset, and the scoring results are shown in Table.~\ref{result5}.
Please refer to the work SummEval~\cite{fabbri2021summeval} for a detailed introduction to these mainstream summarization models.
We bold the top 5 scores of each quality dimension for further experimental analysis.
We find that ROUGE favors the abstractive models, but SummScore seems to prefer the extractive models.
In particular, the LEAD-3 model has achieved high SummScore scores on all four fine-grained qualities.
The reason is that the first three sentences of the news are the most important part of the full text and the LEAD-3 is very suitable for news data.
For Fluency and Consistency, extractive models tend to achieve higher scores.
This is reasonable, because the summary of the abstractive model is generated according to the probability distribution of words, and the problems of fragment repetition and syntax errors can not be avoided.
The generated summary may also have illusory facts that are inconsistent with the facts of the original text.
However, the extractive model produces summaries by splicing sentences extracted from the original text.
Because the sentences are written manually, this avoids grammatical errors and repetition.
Moreover, the sentences are derived from the original, so there is no illusory fact.
For Coherence and Relevance, there is a strong correlation between the two scores.
Moreover, the ROUGE score is also correlated with the score of SummScore in these two quality dimensions.
Almost models with high ROUGE scores also have high scores of Coherence and Relevance and vice versa.

\section{Conclusion}

In this paper, we propose SummScore based on the semantic matching principle of information retrieval, which is a trained scoring metric for summary quality evaluation based on Cross-Encoder.
SummScore has good interpretability. 
It consists of four submodels, which measures the quality of the summary comprehensively from four fine-grained qualities of Coherence, Consistency, Fluency, and Relevance.
We use semi-supervised multi-round training to improve model performance on limited annotated data.
Extensive experiments show that SummScore significantly outperforms existing metrics in terms of human relevancy and helps improve the diversity of generated summaries.
Finally, we also use SummScore to evaluate 16 mainstream summarization models and publish the results for later research.

\bibliographystyle{splncs04}
\bibliography{ref}

\end{document}